\documentclass[a4paper,twoside]{article}

\usepackage{epsfig}
\usepackage{subcaption}
\usepackage{calc}
\usepackage{graphicx}
\usepackage{amssymb}
\usepackage{amstext}
\usepackage{amsmath}
\usepackage{amsthm}
\usepackage{multicol}
\usepackage{pslatex}
\usepackage{apalike}
\usepackage{algorithm2e}
\usepackage[bottom]{footmisc}
\usepackage{hyperref}
\usepackage{epstopdf}
\usepackage{SCITEPRESS}     
\usepackage{color}

\begin{document}

\title{Precision Aquaculture: An Integrated Computer Vision and IoT Approach for Optimized Tilapia Feeding}

\author{\authorname{Rania Hossam\sup{2}, Ahmed Heakl \sup{1} and Walid Gomaa \sup{1,}\sup{3}}
\affiliation{\sup{1}Department of Computer Science and Engineering, Egypt-Japan University of Science and Technology, Alexandria, Egypt}
\affiliation{\sup{2}Computer Science Faculty, Mansoura University, Mansoura, Egypt}
\affiliation{\sup{3}Faculty of Engineering, Alexandria University, Alexandria, Egypt}
\email{raniaelbadry@std.mans.edu.eg,\{ahmed.heakl, walid.gomaa\}@ejust.edu.eg}
}


\abstract{Traditional fish farming practices often lead to inefficient feeding, resulting in environmental issues and reduced productivity. We developed an innovative system combining computer vision and IoT technologies for precise Tilapia feeding. Our solution uses real-time IoT sensors to monitor water quality parameters and computer vision algorithms to analyze fish size and count, determining optimal feed amounts. A mobile app enables remote monitoring and control. We utilized YOLOv8 for keypoint detection to measure Tilapia weight from length, achieving 94\% precision on 3,500 annotated images. Pixel-based measurements were converted to centimeters using depth estimation for accurate feeding calculations. Our method, with data collection mirroring inference conditions, significantly improved results. Preliminary estimates suggest this approach could increase production up to 58 times compared to traditional farms. Our models, code, and dataset are open-source~\footnote{Models: \color{blue}{\url{huggingface.co/Raniahossam33/fish-feeding}}}\footnote{Datasets: \color{blue}{\url{huggingface.co/datasets/Raniahossam33/fish_feeding}}}\footnote{Code: \color{blue}{\url{https://github.com/ahmedheakl/fish-counting}}}.
}

\onecolumn \maketitle \normalsize \setcounter{footnote}{0} \vfill

\section{\uppercase{Introduction}}\label{sec:introduction}

\par 
Optimizing the fish feeding process is critical, as it accounts for up to 40\% of total production costs~\cite{atoum2014automatic,arditya2021design,oostlander2020microalgae}. Effective nutrient control enhances profitability in aquaculture by preventing waste and maintaining high fish quality. Nutrient wastage not only escalates costs but also contributes to water pollution, adversely affecting fish survival and fertility rates. Therefore, precise nutrient management is essential for both economic efficiency and sustainable aquaculture development, ensuring optimal water quality and operational success.

\par 
Recent research has proposed various techniques for controlling the amount of nutrients given to fish. Some researchers have utilized Convolutional Neural Networks (CNNs) for predicting morphological characteristics such as overall length and body size by detecting keypoints on the fish body~\cite{su2009survey,tseng2020automatic}. For instance, \cite{tseng2020automatic} proposed a CNN classifier to detect only two keypoints, the fish head, and tail fork regions, to measure the fish body length. Alternatively, \cite{su2009survey} used a combination of a faster R-CNN~\cite{ren2015faster} for initial fish detection and a stacked hourglass~\cite{newell2016stacked} for keypoint detection, resulting in a complex and computationally expensive method. Another study~\cite{li2021marine} proposed a CNN for marine animal segmentation, which performed well but involved 207.5 million trainable parameters, making it unsuitable for resource-constrained environments like embedded systems or mobile devices.

\par 
Automatic control of fish feeding in real environments remains challenging due to variable data appearance and weather conditions, which can affect the accuracy of detection and tracking results~\cite{soetedjo2019improving,vaquero2021siammt,babaee2019dual}. Object tracking, an active research area in computer vision applications, faces increased complexity in multiple-object tracking due to the need for accurate association of objects across frames~\cite{vaquero2021siammt,tang2017multiple,zhang2020simple}. Recent advancements like the SiamRPN tracker~\cite{zhu2018distractor,li2019siamrpn++} and multi-aspect-ratio anchors have significantly improved the performance of Siamese-network-based trackers by addressing the bounding box estimation problem.

\par 
To address these challenges, our approach involves the following contributions:
\begin{itemize}
    \item We develop a method to estimate the weight of Tilapia fish using a length-weight relationship
    \item We curate an open-source dataset of Tilapia fish images, annotated with keypoints such as the mouth, peduncle, belly, and back.
    \item We train a YOLOv8 model on this dataset, achieving high precision and recall in keypoint detection and fish counting.
    \item We design an end-to-end system, powered by two cameras installed in the fish tank, to monitor feeding amounts, pH levels, and dissolved oxygen. The collected data is relayed to a mobile application for easy access and real-time monitoring.
\end{itemize}

\par 
This approach provides a holistic solution for efficient and effective aquaculture management. 

\par 
The remainder of the paper is organized as follows. Section~\ref{related_work} reviews existing fish mass estimation techniques and feeding methods. Section~\ref{methods} details our approach, including data collection, model training, and fish weight estimation. Section~\ref{iot_system} describes our IoT system architecture for real-time monitoring and control. Results and comparative analysis are presented in Section~\ref{results}, followed by a discussion of limitations and future work in Section~\ref{limitations}. Finally, Section~\ref{conclusion} summarizes our findings and their implications for aquaculture productivity.

\section{\uppercase{Related Work}}\label{related_work}

\par 
This section provides an overview of existing research relevant to our study on precise fish feeding in aquaculture. We focus on two key areas: fish mass and length estimation techniques, and automated fish feeding methods. By examining current approaches, we aim to contextualize our work within the field and highlight the advancements offered by our proposed system.

\subsection{Fish Mass and Length Estimation Techniques}\label{estimation_methods}

\par 
The authors in~\cite{zhang2020estimation} developed a fish mass estimation approach by constructing an experimental data collection platform to capture fish images. 
They used the GrabCut algorithm~\cite{rother2004grabcut} for image segmentation, followed by image enhancement and binarization to extract fish body contours. Shape features were extracted and redundant features were removed using Principal Component Analysis (PCA)~\cite{mackiewicz1993principal}, with feature values calculated through a CF-based (Collaborative Filtering) method~\cite{su2009survey}. A BPNN algorithm was then employed to construct the fish mass estimation model. In contrast, our study uses the YOLOv8 model for keypoint detection to identify critical points such as the fish's head and tail. Our dataset is collected using dual-synchronous orthogonal network cameras, with frames analyzed by our backend server. Instead of traditional feature extraction and PCA, we integrate depth estimation using the GLPN~\cite{depth-estimation} model to create depth maps, enhancing length measurement accuracy by converting pixel coordinates to real-world dimensions. 

\par 
The authors in~\cite{jisr2018length,mathiassen2011high,islamadina2018estimating} have used computer vision like saliency map, edge detection and thresholding and traditional image processing techniques like noise reduction and contrast enhancement to segment~\cite{jisr2018length,islamadina2018estimating}, or make a 3D model~\cite{mathiassen2011high} of the fish body; then use classical machine learning methods, e.g. regression~\cite{sanchez2018automatic,mathiassen2011high} for weight and length extraction. Although these studies have achieved significant results, they have involved complex image processing and feature engineering processes to suit their experimental conditions.

\par 
The authors in~\cite{saleh2023mfld} applied a novel end-to-end keypoint estimation model called MFLD-net. It builds upon CNNs~\cite{sandler2019non}, vision transformers~\cite{dosovitskiy2020image}, and multi-layer perceptrons (MLP-Mixer)~\cite{tolstikhin2021mlp}. Additionally, it leverages patch embedding~\cite{dosovitskiy2020image}, and spatial/channel locations mixing~\cite{tolstikhin2021mlp}. It differs significantly from our approach as the images were taken outside the pool environment. Additionally, their method involved annotating more than four keypoints on each fish, which may increase the complexity of the annotation process. Furthermore, their study used fish of a fixed size, which limits the model's ability to generalize to different fish sizes.

\subsection{Fish Feeding Techniques}\label{feeding_techniques}

\par 
The authors in~\cite{riyandani2023computer} focus on developing an automatic feeder employing the YOLOv5x detection model~\cite{vasanthi2024multi} for fish feed detection. Their model achieved notable metrics, including an accuracy of 82\% and mAP of 81.9\%. The automatic feeder dispenses a fixed amount of 30 grams of fish feed every five rotations of the stepper motor, with observed variations in fish feed consumption patterns throughout the day. In contrast, our research advances this field by utilizing the more advanced YOLOv8 model for keypoint detection, which promises improved performance. This allows us to adjust the feeding amount dynamically based on real-time measurements of each fish's size, optimizing feeding practices and preventing overfeeding. Our approach integrates depth estimation to convert 2D image measurements into real-world dimensions, enhancing the accuracy of fish length and weight estimation, and consequently, the daily feeding allowance. This provides a more precise, scalable, and tailored feeding mechanism than the fixed feeding amount used in their study.

\par 
The authors in~\cite{tengtrairat2022non} employ a Mask R-CNN~\cite{He_2017_ICCV} with transfer learning to detect Tilapia fish in images. The detection model identifies the fish and extracts dimensions such as length and width. The subsequent weight estimation relies on regression learning models utilizing three key features: fish length, width, and depth. The researchers investigated a regression method for weight estimation like support vector regression~\cite{awad2015support}. However, the methodology has several cons. Despite its accuracy, the use of Mask R-CNN is computationally intensive and requires significant processing power. The multi-step process involving depth estimation and feature extraction increases complexity and can be error-prone. Additionally, the regression models, while effective, require precise input features and may not generalize well across varying conditions.

\begin{figure}[t]
    \centering
    \includegraphics[width=0.48\textwidth]{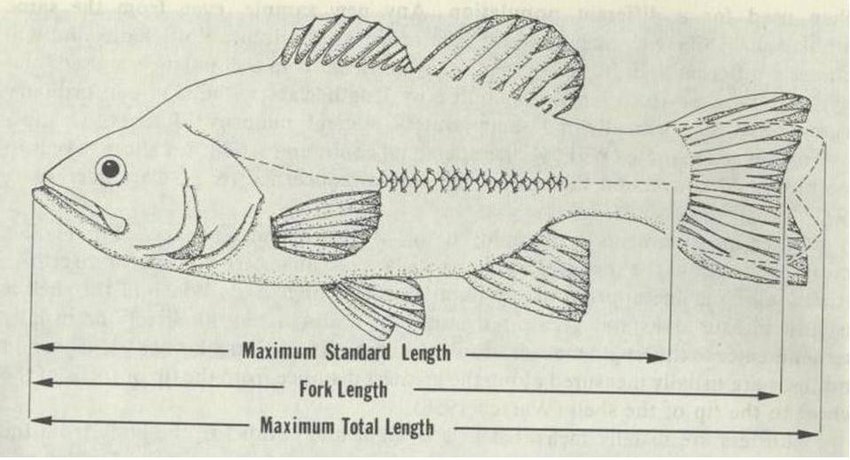}
    \caption{Description of fish lengths. We opt to measure the maximum standard length to calculate fish weight~\cite{froese2014length}.}
    \label{fig:dis_label}
\end{figure}

\section{\uppercase{Methods}}\label{methods}

\par 
This section details our approach to developing a precise fish-feeding system. We describe the process of estimating Tilapia fish weight, our data collection and annotation methods, the implementation of YOLOv8 for keypoint detection and fish counting, and our technique for calculating fish length and feed amounts. These methods form the foundation of our integrated computer vision and IoT-based solution for optimizing aquaculture management.

\begin{figure}[t]
    \centering
    \subfloat[Keypoint annotation.]{%
        \includegraphics[width=0.2\textwidth]{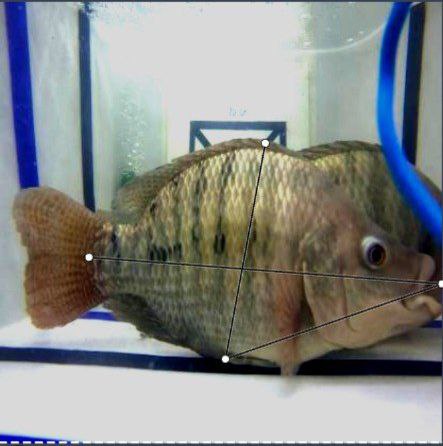} 
        \label{key_tilapia}
    }\hfill
    \subfloat[Detection example.]{%
        \includegraphics[width=0.2\textwidth]{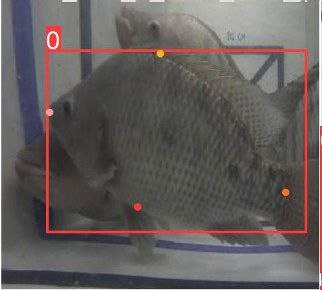}
        \label{fig:tilapia-counting}
    }
    \caption{Examples for annotation for fish counting.}
    \label{fig:tilapia_examples}
\end{figure}

\subsection{Tilapia Fish Weight Estimation}

\par 
To determine the appropriate amount of Tilapia fish feed,~\cite{jisr2018length,tilapia_feed_management} have shown that the amount of feed required for fish can be estimated based on their weight. \cite{m2020fisheries} provided an equation~\ref{eq:weight_len} to estimate the fish weight from its length, where the length is defined as the distance from the mouth to the peduncle~\cite{jerry1998morphological} as shown in figure \ref{fig:dis_label} (Maximum Standard Length):

\begin{equation}
W = aL^b
\label{eq:weight_len}
\end{equation}

where $W$ is the fish's weight in grams (g), $L$ is the length (cm), and $a$ and $b$ are species-specific coefficients ($a = 0.014$ and $b = 3.02$ for Tilapia)~\cite{m2020fisheries}. This method simplifies data collection, bypassing direct weight measurements, and allows weight distribution analysis and other parameters within the fish population.

\subsection{Keypoint Annotation and Data Collection}

\par 
To determine the appropriate feed amount for Tilapia fish, we first needed to specify keypoints to measure their length accurately, defined as the distance from the mouth to the peduncle~\cite{example_fish_length}. For this purpose, we collected 3,500 images of Tilapia fish in a small bowl of three fish. These images were manually annotated using Roboflow~\cite{roboflow}, a widely used tool for creating and managing annotated datasets. Although we only needed the mouth and peduncle keypoints, we annotated four keypoints on each fish—mouth, peduncle, belly, and back—to aid future research using girth to determine weight (see figure~\ref{key_tilapia}). Following the annotation process, we trained YOLOv8 model~\cite{reis2023real} on the respective dataset to predict the keypoints accurately.
 
\subsection{Calculating Fish Length}

\par 
To estimate fish length, we calculate the Euclidean distance between the head and tail keypoints in pixel units. This measurement is converted to centimeters by integrating depth estimation and focal length, considering the camera distance. Initially, each fish image is resized to a standard ($416\times416$) for consistency.

\begin{table}[h]
    \centering
    \caption{Daily feeding allowances as a percentage of fish weight \cite{riche2003feeding}}.
    \resizebox{0.49\textwidth}{!}{
    \begin{tabular}{|c|c|}
        \hline
        \textbf{Fish Weight Range (g)} & \textbf{Daily Feeding Range (\%)} \\
        \hline
        0--1   & 10 to 30 \\
        1--5   & 6 to 10  \\
        5--20  & 4 to 6   \\
        20--100& 3 to 4   \\
        Larger than 100 & 1.5 to 3 \\
        \hline
    \end{tabular}
    }
    \label{tab:feeding-allowances}
    \small
\end{table}

\par 
The GLPN (Global-Local Path Networks) model~\cite{depth-estimation} is employed for depth estimation, predicting a depth value for each pixel and creating a depth map essential for spatial information. The YOLOv8 model detects keypoints on the fish (head and tail). These coordinates are then adjusted by their respective depth values to approximate real-world distances. In this depth-adjusted coordinate space, the Euclidean distance between the head and tail keypoints represents the fish’s length in pixel units.
For the conversion of pixel coordinates to real-world coordinates, given a point in the image with coordinates $(x_p, y_p)$ and depth $d$, the real-world coordinates $(X, Y, Z)$ can be computed as follows:
Let $f$ be the focal length of the camera (in pixels). Then, we can calculate the 3D coordinates $(X, Y, Z)$ from the 2D image coordinates $(x_p, y_p)$ and depth $d$ as follows:

\begin{equation}
X = \frac{x_p \cdot d}{f},\quad Y = \frac{y_p \cdot d}{f},\quad  Z = d
\end{equation}

\begin{figure*}
    \centering
    \includegraphics[width=0.9\linewidth]{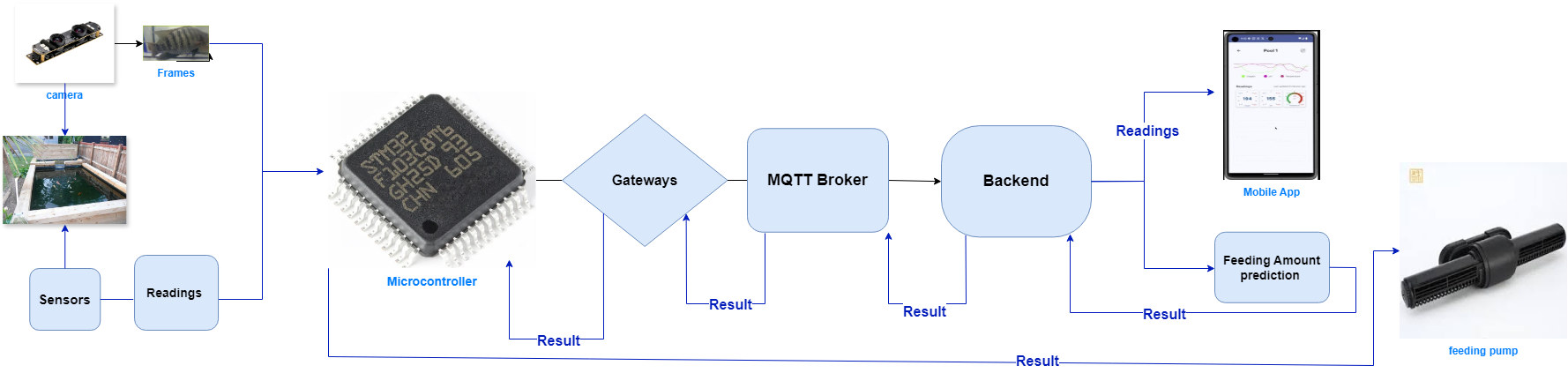}
    \caption{Our IoT system architecture and flow for automated aquarium monitoring and feeding.}
    \label{fig:IOT_ARCH}
\end{figure*}

\par 
We can retrieve the length by getting the Euclidean distance between $(X, Y, Z)_{head}$ and $(X, Y, Z)_{tail}$ which we call $distance$. Finally, the fish length is calculated using the formula:
\begin{equation}
\text{fish length} = \frac{f}{\text{distance}}
\label{actual_dis}
\end{equation}

\subsection{Calculating Fish Count}

\par 
After estimating the fish feed amount based on weight, our next goal is to determine the total feed required for the fish in the bowl. To achieve this, we trained another YOLOv8 model~\cite{reis2023real} on our dataset to count the fish accurately as shown in figure~\ref{fig:tilapia-counting}.

\subsection{Feed Estimation}

\par 
Once the optimal fish feeding allowances are determined from table~\ref{tab:feeding-allowances}, we estimate the final feed requirements. By leveraging the robustness of the fish counting models in table~\ref{tab:metrics_comparison}, the final feed estimation is calculated by multiplying the number of fish, as determined by the fish counting model, with the average feeding amount.

\section{\uppercase{IoT System}}\label{iot_system}
\begin{figure}[h]
    \centering
    \includegraphics[width=0.40\textwidth]{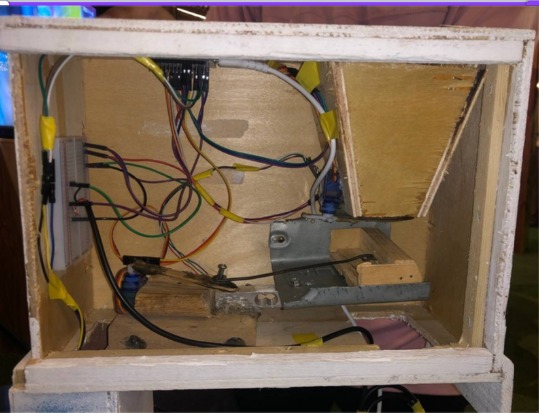}
    \caption{Interior view of the prototype fish feeding system, showing electrical wiring, sensors, and mechanical components housed within a wooden enclosure.}
    \label{fig:protoype}
\end{figure}

\par 
Our IoT system integrates a diverse set of sensors including pH, dissolved oxygen (DO), and temperature sensors, along with two cameras, an STM32F103C8 MicroController Unit (MCU)~\cite{stm32f103c8}, and dual pumps—one for feeding fish and another for pH control, as depicted in figure~\ref{fig:IOT_ARCH}. These sensors are connected to the MCU and continuously collect crucial data from the aquatic environment. A prototype is shown in figure~\ref{fig:protoype}.

\par 
The sensor readings are initially processed by the MCU. Once processed, the MCU
transmits the data to gateways within the system architecture. From the gateways, the data is then forwarded via the MQTT communication protocol~\cite{mqtt} to our backend server. The backend server then acts as the central hub where the data is stored and processed.

\par
\begin{figure}[h]
    \centering
    \includegraphics[width=0.35\textwidth]{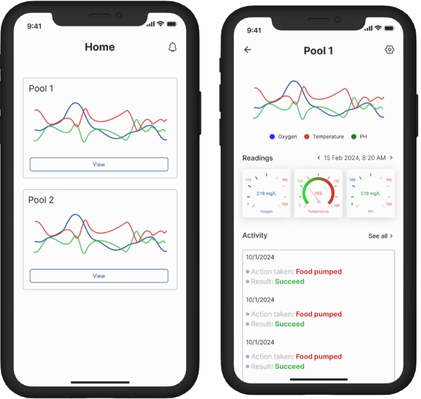}
    \caption{Mobile app readings.}
    \label{fig:mobile}
\end{figure}

 
The backend server interacts with a dedicated mobile application, shown in figure~\ref{fig:mobile}, serving as the user interface. Through this application, users can view real-time graphs, detailed analytics, statistical summaries, and logs reflecting the system's operations and environmental conditions. Simultaneously with sensor data collection, our dual-synchronous orthogonal network cameras actively capture frames from the pool. These frames undergo processing via the MCU, followed by transmission to the gateway, and onward to the MQTT broker before reaching the backend server. At the backend, AI models analyze these frames from two cameras to extract keypoints and fish counts. The results from each camera data and averaged for more accurate predictions.

\begin{figure}[b]
    \centering
    \includegraphics[width=0.45\textwidth]{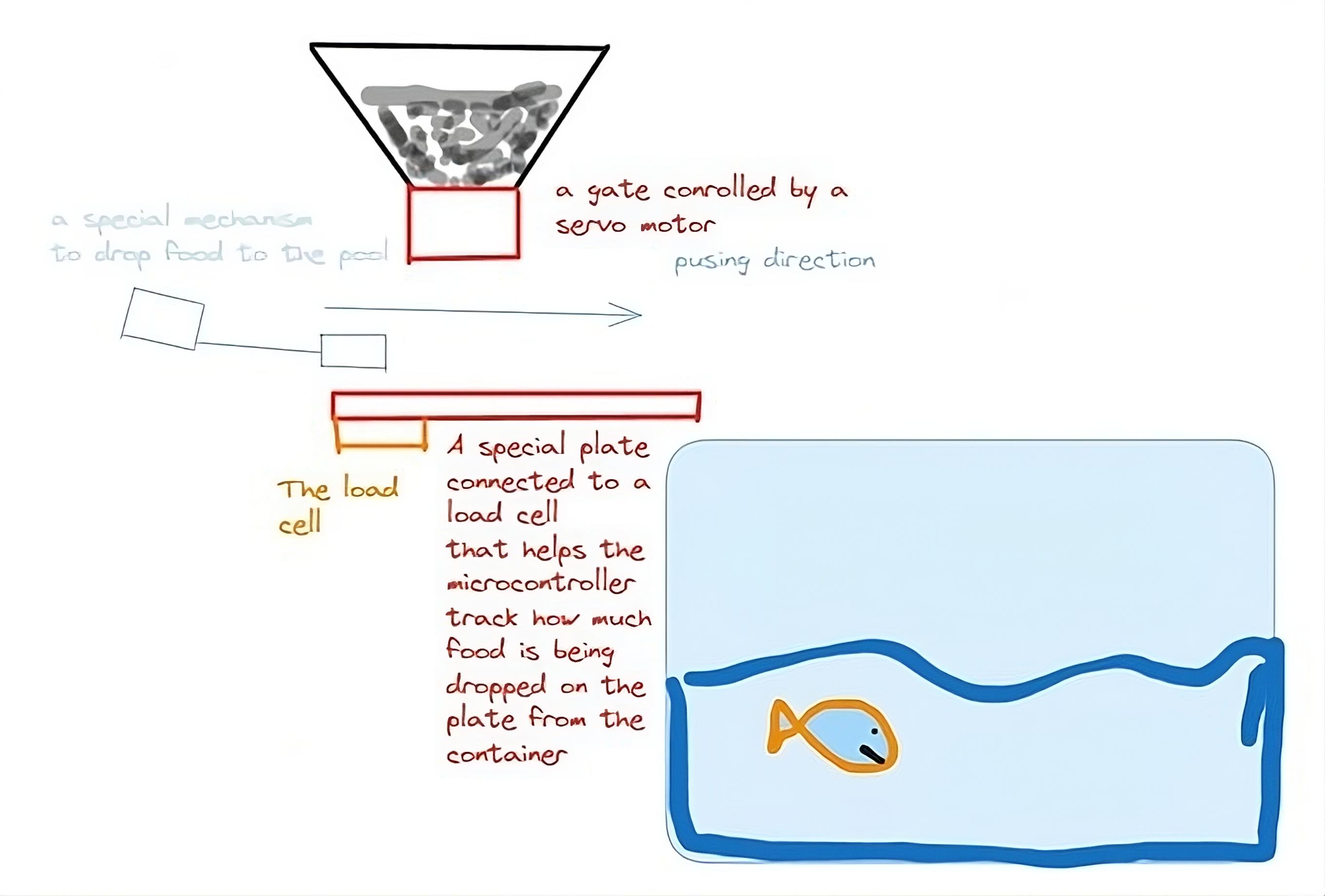}
    \caption{Feeding pump mechanism.}
    \label{fig:feeding}
\end{figure}

\par 
The AI model's predictions are then relayed back to the backend server, which communicates them via the MQTT broker to the MCU. Based on these insights, the MCU precisely transmits the feeding amount to the feeding pump mechanism.
The feeding mechanism operates through a vertical inventory above the pool, regulated by gates, and monitored by a load cell sensor for precise food dispensation. This setup includes a 10KG load cell, an HX-711 amplifier, and two servo motors for meticulous gate control as shown in figure~\ref{fig:feeding}.

\section{\uppercase{Results \& Discussion}}
\label{results}

\par
This section presents the outcomes of our experiments using the YOLOv8 model for keypoint detection and fish counting in Tilapia aquaculture. Our evaluation demonstrates that YOLOv8 outperforms existing approaches in both accuracy and speed, making it exceptionally well-suited for deployment on edge devices in resource-constrained environments.

\subsection{YOLOv8 Performance on Tilapia Dataset}

\par 
Our experiments demonstrate the superior performance of our YOLOv8 model in both keypoint detection and fish counting tasks. Table \ref{tab:metrics_comparison} summarizes the evaluation metrics for the YOLOv8 model trained on our custom Tilapia fish dataset. 

\begin{table}[h]
\centering
\caption{Evaluation metrics for keypoints detection and fish-counting models trained on Tilapia fish dataset.}
\resizebox{0.49\textwidth}{!}{
    \begin{tabular}{|l|l|l|l|l|}
    \hline
    \textbf{Method} & \textbf{Precision} & \textbf{Recall} & \textbf{AP@50} & \textbf{ AP@75}\\
    \hline
    YOLOv8 Keypoints & 94.50 & 89.71 & 99.68 & 94.16 \\
    \hline
    YOLOv8 Counting & 96.21 & 86.82 & 98.88 & 92.47 \\
    \hline
    \end{tabular}
}
\label{tab:metrics_comparison}
\end{table}

\begin{table*}[t]
\centering
\caption{Comparison of various scores across different models on the Tilapia dataset on fish Keypoints detection.}
\begin{tabular}{|l|c|c|c|c|c|}
\hline
\textbf{Method} & \textbf{Precision} & \textbf{Recall} & \textbf{AP@50} & \textbf{AP@75} & \textbf{AP}\\
\hline
Faster R-CNN~\cite{ren2015faster} & 91.72 & 85.99 & 98.50 & 90.19 & 67.04 \\ 
\hline
Mask R-CNN~\cite{He_2017_ICCV}& 92.61 & 87.34 & 99.11 & 92.12 & \textbf{75.68} \\ 
\hline
RetinaNet~\cite{lin2017focal} & 90.79 & 84.26 & 98.17 & 83.56 & 60.53 \\ 
\hline \hline
\textbf{YOLOv8 (Ours)} & \textbf{94.96} & \textbf{89.06} & \textbf{99.68} & \textbf{94.16} & 68.04 \\ 
\hline
\end{tabular}
\label{tab:comparison}
\end{table*}

\par
For the fish counting task, we employed the same YOLOv8 architecture, which yielded an even higher precision of 96.21\%. This precision metric, derived from the model's training report, represents the ratio of correctly detected fish to the total number of detections made by the model. Our fish counting method utilizes frames captured simultaneously from two cameras positioned at different angles in the fish farm. The system processes these paired frames to provide a more comprehensive view of the fish population, helping to reduce occlusions and improve counting accuracy.
\par
To validate these results and assess real-world performance, we conducted additional tests on a separate dataset of 100 frame pairs (200 images total) from actual fish farm conditions. In these tests, the model achieved a counting accuracy of 94.5\%, with an average absolute error of 0.7 fish per frame pair. This close alignment between training metrics and real-world performance underscores the model's reliability in practical applications.
\par
The high precision and real-world accuracy of our fish counting model are crucial for accurate population estimation, which directly impacts feed calculation. By combining this accurate count with the average feeding amount determined from our feeding allowance table, we can achieve precise feed estimation, minimizing overfeeding and reducing both feed waste and potential water pollution.

\subsection{Comparative Analysis with Existing Models}

\par 
To contextualize our results, we compared YOLOv8's performance with other state-of-the-art deep learning models, as reported in \cite{tengtrairat2022non}. Table~\ref{tab:comparison} presents the comparison of various metrics across different models (Faster R-CNN\cite{ren2015faster}, Mask R-CNNN~\cite{He_2017_ICCV}, RetinaNet~\cite{lin2017focal}, and our YOLOv8 models~\cite{vasanthi2024multi}) on the tilapia dataset.

\par
As evident from table \ref{tab:comparison}, our YOLOv8 model outperforms other models in most metrics, particularly in AP@50 and AP@75. The superior performance in AP@75 is especially noteworthy, as it indicates YOLOv8's ability to maintain high accuracy even with stricter overlap requirements. This is crucial for precise keypoint detection in densely populated fish farms.

\par 
While Mask R-CNN shows a higher overall AP score, which averages performance across all IoU thresholds, YOLOv8 demonstrates more consistent performance at the critical AP@50 and AP@75 levels. This suggests that YOLOv8 may be more reliable for practical applications where moderate to high precision is required.

\subsection{YOLOv8 \& Edge Computing}

\par
YOLOv8's architecture is optimized for edge computing, making it ideal for real-time aquaculture monitoring. Its lightweight design allows efficient processing on limited-resource devices, with the nano version achieving sub-200ms inference times on our MCU. This efficiency reduces energy consumption and operational costs. YOLOv8's scalability ensures consistent performance across various hardware configurations, from IoT devices to edge servers. Local data processing minimizes latency and enables rapid decision-making without constant server communication. These features address on-site aquaculture management challenges, potentially revolutionizing Tilapia monitoring. YOLOv8's reliable performance under hardware constraints makes it a superior choice for transforming aquaculture practices.

\subsection{Implications for Aquaculture Productivity}

\par 
The high accuracy of our YOLOv8-based system translates to significant potential improvements in aquaculture productivity. Based on preliminary assessments and comparisons with traditional methods, we estimate that our approach can contribute to a 58-fold increase in production compared to conventional fish farms, inspired by \cite{seafdec2022}. This dramatic improvement is attributed to:
\begin{enumerate}
    \item More accurate fish counting, enabling optimal stocking densities.
    \item Precise monitoring of fish growth and health through keypoint detection.
    \item Reduced water pollution and fish mortality due to timely interventions.
\end{enumerate}

\par 
It is important to note that these productivity gains are theoretical maximums based on optimal conditions and full implementation of our system. Real-world results may vary depending on specific farm conditions and management practices.

\section{\uppercase{Limitations and Future Work}}\label{limitations}

\par 
The limitations of this study include the use of datasets from a single fish size in a controlled environment. Future work should include a diverse range of fish sizes and environments to improve model generalizability, especially for smaller fish where keypoint detection is more challenging. Additionally, the current system does not account for varying environmental factors such as water quality, which can influence fish growth and feeding behavior. Integrating environmental monitoring could further optimize feeding practices. While the YOLOv8 model performed well on the Tilapia dataset, its applicability to other fish species remains untested. Lastly, expanding training datasets to include multiple species could enhance its utility across different aquaculture contexts. 

\section{\uppercase{Conclusion}}\label{conclusion}

\par 
This paper used computer vision and IoT technologies to present a novel system for precise Tilapia fish feeding. The system utilizes real-time water quality monitoring and vision-based fish weight estimation to determine optimal feeding amounts. Our models demonstrated superior performance with precision of \textbf{94\%} for keypoint detection, and \textbf{96\%} for fish counting, respectively, outperforming Faster R-CNN, Mask R-CNN, and RetinaNet in key metrics.  This study provides a precise, scalable solution for sustainable and efficient aquaculture, with recommendations for further real-world testing and refinement. Lastly, this approach has the potential to significantly enhance fish farm productivity (up to 58x) while mitigating environmental concerns by minimizing pollution and fish mortality.

\section{\uppercase{Acknowledgements}}
This work would not have been possible without the valuable contributions of our collaborators. We extend our sincere gratitude to Menna Eid and Mohamed Salman from Mansoura University for their expertise and assistance in developing the IoT system and mobile application. Their contributions were instrumental in bringing this project to fruition.
    
\bibliographystyle{apalike} 
{\small
\bibliography{example}}

\end{document}